\documentclass[authoryear, hidelinks]{article}

\usepackage[english]{babel}

\usepackage[letterpaper,top=2cm,bottom=2cm,left=3cm,right=3cm,marginparwidth=1.75cm]{geometry}

\usepackage[utf8]{inputenc} 	
\usepackage{graphicx}      		
\usepackage[numbers]{natbib}	
\usepackage{amsfonts}        	
\usepackage{amsmath,amssymb}    
\usepackage{bm}					
\usepackage{xcolor}             
\usepackage{float}              
\usepackage[caption=false]{subfig} 
\usepackage{multirow}  			
\usepackage{tikz}				
\usetikzlibrary{decorations.pathreplacing,calligraphy}
\usepackage{hyperref}           

\usepackage{algorithm}
\usepackage{algpseudocode}

\usepackage{authblk}            

\newcommand{\figasset}{
	\begin{tikzpicture}
	
	\def\MAX{7.0}
	\def\UNIT{0.1*\MAX}
	\def\HALF{0.5*\UNIT}
	\def\QUART{0.5*\HALF}
	
	\def\VA{\HALF}
	\def\VB{\VA + \HALF}
	\def\VC{\VA + \UNIT}
	\def\VW{0.2*\UNIT}
	
	\def\YPIPE{2.5*\UNIT}
	
	\def\XSEP{7*\UNIT}
	\def\SW{1.0*\UNIT}
	\def\SR{0.5*\UNIT}
	\def\XSB{\XSEP+\SR}
	\def\XSC{\XSB+\SW}
	\def\XSD{\XSC+\SR}
	\def\XSM{\XSB+0.5*\SW}
	\def\XQ{\XSD+\HALF}
	
	\def\YQW{\YPIPE-\SR-\QUART}
	\def\YQG{\YPIPE+\SR+\QUART}
	
	\foreach \x/\l in {1, 2, 3, 5/m}{
		
		\draw [thick, dashed] (\UNIT*\x, 0) -- (\UNIT*\x, -\HALF);
		\draw [thick, ] (\UNIT*\x, 0) -- (\UNIT*\x, \VA);
		\draw [thick, ->] (\UNIT*\x, \VC) -- (\UNIT*\x, \YPIPE);
		\draw [thick, ->] (\UNIT*\x-\VW, \VA) -- (\UNIT*\x+\VW, \VA) -- (\UNIT*\x-\VW, \VC) -- (\UNIT*\x+\VW, \VC) -- cycle;
		
		\node[anchor=west] at (\UNIT*\x, \VB) {$y_{\l}$};
	}
	\draw [thick, dotted] (\UNIT*3.5, \VB) -- (\UNIT*4.5, \VB);
	
	\draw [thick, ->] (\UNIT, \YPIPE) -- (\XSEP, \YPIPE);
	
	\draw [thick] (\XSB, \YPIPE-\SR) arc (270:90:\SR cm);
	\draw [thick] (\XSB, \YPIPE-\SR) -- (\XSC, \YPIPE-\SR);
	\draw [thick] (\XSB, \YPIPE+\SR) -- (\XSC, \YPIPE+\SR);
	\draw [thick] (\XSC, \YPIPE-\SR) arc (-90:90:\SR cm);
	
	\draw [thick] (\XSC, \YPIPE-\SR) -- (\XSC, \YPIPE+0.25*\SR); 
	
	\draw [thick, ->] (\XSD, \YPIPE) -- (\XQ, \YPIPE) node[anchor=west]{$y_{O}$};
	\draw [thick, ->] (\XSM, \YPIPE+\SR) -- (\XSM, \YQG) -- (\XQ, \YQG) node[anchor=west]{$y_{G}$}; 
	\draw [thick, ->] (\XSM, \YPIPE-\SR) -- (\XSM, \YQW) -- (\XQ, \YQW) node[anchor=west]{$y_{W}$}; 
	
	\end{tikzpicture}
}

\newcommand{\figwell}{
	\begin{tikzpicture}
	
	\def\MAX{6.5}
	\def\HALF{0.05*\MAX}
	\def\UNIT{0.1*\MAX}
	\def\XA{0.2*\MAX}
	\def\XB{0.4*\MAX}
	\def\MID{0.5*\MAX}
	\def\XC{0.6*\MAX}
	\def\XD{0.8*\MAX}
	
	\foreach \x/\l in {\XA / {Upstream}, \XD/ {Downstream}}{
		\draw[fill] (\x, 0) circle (0.05cm);
	}
	
	\draw[thick] (\XB,-\HALF) -- (\XB,\HALF) -- (\XC,-\HALF) -- (\XC,\HALF) -- cycle;
	
	\draw [thick, ->] (\XB-0.5, -0.5) -- node[below,midway]{$y$} (\XC+0.5, -0.5);
	
	\draw [thick, ->] (-0.0, 0) -- (\XB, 0);
	\draw [thick, ->] (\XC, 0) -- (\MAX, 0);
	
	\foreach \x/\l in {\XA / {$p_1, T$}, \MID/ {$u$}, \XD / {$p_2$}}{
		\draw [thick,dashed] (\x, 0) -- (\x, \UNIT);
		\draw[thick] (\x, \UNIT+0.5) circle (0.5cm);
		\node[anchor=center] at (\x, \UNIT+0.5) {\l};
	}

	\node[anchor=center] at (-0.5, \UNIT+0.5) {$\gamma, \lambda$};

	\draw [thick,dashed] (\MAX -0.5, -1.0) -- (\MAX -0.5, 2.0);
	\draw [thick,dashed] (0.5, -1.0) -- (0.5, 2.0);
	
	\node[anchor=east] at (0.5, 2) {Reservoir};
	\node[anchor=west] at (\MAX -0.5, 2) {Facility};	
	\node[anchor=center] at (3, 2) {Well};
		
\end{tikzpicture}
}

\title{Sequential Monte Carlo applied to virtual flow meter calibration}

\author[1,2]{Anders T. Sandnes\footnote{Corresponding author: anders@solutionseeker.no}}
\author[1,3]{Bjarne Grimstad}
\author[2]{Odd Kolbjørnsen}

\affil[1]{Solution Seeker AS, Oslo, Norway}
\affil[2]{Department of Mathematics, University of Oslo, Oslo, Norway}
\affil[3]{Department of Engineering Cybernetics, Norwegian University of Science and Technology, Trondheim, Norway}

\begin{document}
\maketitle

\begin{abstract}
Soft-sensors are gaining popularity due to their ability to provide estimates of key process variables with little intervention required on the asset and at a low cost.
In oil and gas production, virtual flow metering (VFM) is a popular soft-sensor that attempts to estimate multiphase flow rates in real time.
VFMs are based on models, and these models require calibration. 
The calibration is highly dependent on the application, both due to the great diversity of the models, and in the available measurements.
The most accurate calibration is achieved by careful tuning of the VFM parameters to well tests, but this can be work intensive,
and not all wells have frequent well test data available.
This paper presents a calibration method based on the measurement provided by the production separator,
and the assumption that the observed flow should be equal to the sum of flow rates from each individual well.
This allows us to jointly calibrate the VFMs continuously.
The method applies Sequential Monte Carlo (SMC) to infer a tuning factor and the flow composition for each well.
The method is tested on a case with ten wells, using both synthetic and real data.
The results are promising and the method is able to provide reasonable estimates of the parameters without relying on well tests.
However, some challenges are identified and discussed, particularly related to the process noise and how to manage varying data quality.
\end{abstract}

\section{Introduction}

The fourth industrial revolution has brought incredible progress in terms of data acquisition, storage, and availability \citep{Kagermann2022}. 
Large quantities of industrial data are being collected and stored in real-time. 
As the quantity of information grows, human data analysts are increasingly dependent on machine intelligence systems to process and filter data \citep{Kagermann2022}.
Soft sensors are predictive models that run on top of these data streams to provide insight and situational awareness for engineers and operators \citep{KADLEC2009}.
Soft sensors are mathematical models combined with supporting software.
This makes them cost-efficient to install and maintain, as they do not require additional hardware on-site.
The models can be data-driven, based on physics and process knowledge, or a combination of both methodologies.

In oil and gas production, the data infrastructure is well-established,  but the utilization of real-time data is still limited \citep{Lu2019}. 
An application that has seen significant research is Virtual Flow Meters (VFMs) \citep{Bikmukhametov2020}.
A virtual flow meter estimates gas, oil, and water flow rates using a mathematical model and commonly available measurements.
Several modeling strategies are presented in the literature, covering physical models \citep{Schuller2003},
machine learning models \citep{Sandnes2021}, and hybrid models \citep{Hotvedt2022a}.
Virtual flow meters are traditionally based on one or more pressure drops in the system,
such as over a choke valve or a long section of a pipeline.

\begin{figure}
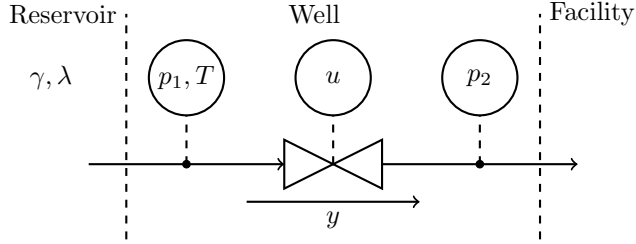

     \centering
	    \figwell
     \caption{Topology of an oil well. Measured is the choke opening $u$, pressures $p_1$ and $p_2$, and temperature $T$. The flow rate $y$ through the well is unknown. The flow originates from a reservoir that has an unknown mixture of gas, oil, and water. The mixture is described by the gas fraction $\gamma$ and the oil factor $\lambda$. }
     \label{fig:well}
 \end{figure}

VFMs are most commonly used to estimate flow from individual wells.
Ideally, the VFM would be able to estimate gas, oil, and water simultaneously,
but the number of measurements available is often insufficient to achieve this. 
Instead, it is common to assume that the flow composition is partially, or fully, known.
This reduces the problem to only infer the total flow rate \citep{Bikmukhametov2020}. 
For instance, consider the example of a well with instrumentation given in Figure \ref{fig:well},
where choke, pressure, and temperature measurements are available continuously in real-time.
It would be challenging to infer all three flow rates using these four measurements.
Instead, \textit{the sum} of gas, oil, and water could be estimated.
This would typically include an assumption about the mixture composition.

Regardless of measurements and modeling methodology, 
the models need to be kept up to date to ensure satisfactory performance.
This is due to changes in the system, which include the flow composition, fluid properties, and operational conditions \citep{Bikmukhametov2020}.
Several methods have been proposed, and they naturally vary significantly based on the type of models and the type of data available for calibration.
\citet{Hotvedt2022} applies online learning to update hybrid and data-driven models based on \textit{well-test data}.
Well-tests are data collected when a well is produced alone on a test-separator,
where the gas, oil, and water rates of the well are directly measured after physical separation. 
Calibration based on well tests is, of course, very precise.
However, because well tests take a significant time to conduct and multiple wells share the same test facility, the frequency of well tests can be limited to a few times a month \citep{Monteiro2020}.

Another calibration strategy is to use a sequence of the continuously available measurements, typically pressures and temperatures,
in combination with state-space methods such as Kalman filters and particle filters \citep{Adukwu2022}.
These methods do not use direct measurements of the flow rates to tune the models.
\citet{LORENTZEN20031283} utilizes an ensemble Kalman filter to calibrate a dynamic model using a series of high-frequency pressure measurements. 
\citet{He2020calibration} applies a particle filter to calibrate a hybrid model with both data-driven and mechanistic model components.

\begin{figure}
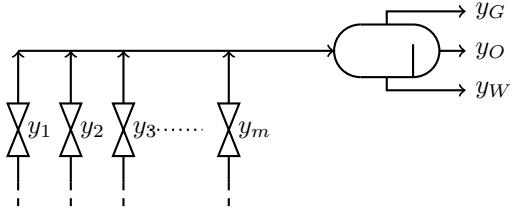

     \centering
	    \figasset
     \caption{Asset with $m$ wells sharing a separator.
     The wells produce to a shared separator, where the flow rates are measured.
     For instance, the gas rate from each well adds up to the separator gas rate, $y_G = \sum_{j=1}^m y_{j,G}$,
     and similarly for oil and water rates.
     }
     \label{fig:asset}
 \end{figure}

A problem similar to VFM calibration is that of rate \textit{allocation} \citep{Kanshio2020}. 
Allocation is concerned with an observed co-mingled flow, which is the sum of production from a collection of wells.
The task is to assign a portion of the observed flow to each of the contributors.
Consider the asset illustrated in Figure \ref{fig:asset}, where $m$ wells produce to a shared separator.
Production allocation would then be to find each well's contribution to the flow rates measured by the separator.
If each well has a virtual flow meter, these flow rates could be used to compute the allocation factors \citep{Kanshio2020}.
This leads to an allocation that respects the mass balance of the system \citep{Badings2020}.

In this paper, we present a VFM calibration approach based on the Sequential Monte Carlo (SMC).
We attempt to estimate the flow composition and tuning factors for a collection of wells using a combination of production data and well tests.
Production data is the joint production from all wells measured after separation, as illustrated in Figure \ref{fig:asset}.
Production data is collected continuously.
The goal is to be able to calibrate the VFMs in the absence of frequent well tests.

The remained of this paper is outlined as follows.
Section \ref{sec:smc} introduces the necessary background on SMC.
Section \ref{sec:problem} describes the problem in detail, 
describing the quantities to be inferred and the available data.
Section \ref{sec:results} presents an empirical study on an asset with ten wells using real and synthetic data.
Section \ref{sec:conclusion} summarizes the results and concludes.

\section{State-space models and Sequential Monte Carlo} \label{sec:smc}

SMC is a class of recursive Bayesian filtering methods \citep{chopin2020introduction}.
While they are applicable to Monte Carlo simulation in general,
they are particularly efficient when applied to infer the latent variables of a Markovian state-space model.
Let $\theta_t$ be the state and $\mathcal{D}_t = \lbrace (y_t, x_t) \rbrace$ be the observations at time step $t$, for $t = 0, 1, \dots, n$.
A Markovian state-space model is then defined by a 
a prior distribution
\begin{align}
p(\theta_0), \label{eq:state-space-prior}
\end{align}
a transition distribution
\begin{align}
    p(\theta_t | \theta_{t-1}), \label{eq:state-space-transition}
\end{align}
which is valid for $t > 0$, and the likelihood of the observations 
\begin{align}
    p(y_t | \theta_t, x_t). \label{eq:state-space-likelihood}
\end{align}
The filtering problem of interest is the posterior $p(\theta_t | \mathcal{D}_{0:t})$,
where the notation $\mathcal{D}_{0:t}$ refer to the set of elements $\lbrace \mathcal{D}_{0}, \dots \mathcal{D}_{t}\rbrace$.
Including the full sequence of states, $\theta_{0:t}$,
the filtering problem can be written recursively as 
\begin{align}
p(\theta_{0:t} | \mathcal{D}_{0:t}) =
\frac{p(y_t | \theta_t, x_t)p(\theta_t | \theta_{t-1})}{p(\mathcal{D}_{t}|\mathcal{D}_{0:t-1})}
p(\theta_{0:t-1} | \mathcal{D}_{0:t-1}), \label{eq:bayes-recursive}
\end{align}
which can be exploited to efficiently update the posterior when new data arrives \citep{doucet2001sequential}.

If the state-space model components in Equation \eqref{eq:state-space-prior} to \eqref{eq:state-space-likelihood} 
are all linear with additive Gaussian noise, 
the filter can be derived analytically as a closed-form solution.
This is the well-known \textit{Kalman filter} \citep{doucet2001sequential}.
For most problems, closed-form solutions do not exist,
and the posterior distributions must be approximated. 
Monte Carlo methods achieve this by generating samples from the posterior distribution and then using these samples to approximate the quantities of interest.

SMC is a sampling scheme that allows observations to be processed sequentially as they appear. 
It is based on importance sampling, 
where samples are drawn from a \textit{proposal distribution} and are weighted according to a true target distribution \citep{doucet2001sequential}.
We apply the version of Sequential Monte Carlo described in Algorithm \ref{alg:guided-particle-filter}.
We refer to \citet{chopin2020introduction} for an in-depth treatment of the algorithm and its variations.

\begin{algorithm}
\caption{Guided particle filter. The superscripts indicate particle number.
All operations must be done for each particle, $i = 1,\dots, N$.}
\label{alg:guided-particle-filter}
\begin{algorithmic}
\State $\theta_0^{(i)} \sim q(\theta_0 | y_0, x_0)$ \Comment{Sample from proposal prior.}
\State $w_0^{(i)} = \frac{p(y_0 | \theta_0, x_0) p(\theta_0)}{q(\theta_0 | y_0, x_0)}$ \Comment{Compute weights.}
\State $W_0^{(i)} = w_0^{(i)} / \sum_{k=1}^N w_0^{(k)}$ \Comment{Normalize weights.}
\For{$t = 1, 2, \dots $}
\State $A_t^{1:N} \gets \text{resample}(W_{t-1}^{(i)})$ \Comment{Resample at every step.}
\State $\theta_t^{(i)} \sim q(\theta_t| \theta_{t-1}^{(A_t^{t})}, y_t, x_t)$ \Comment{Sample from transition proposal.}
\State $w_t^{(i)} = 
\frac{p(y_t | \theta_t, x_t) p(\theta_t | \theta_{t-1}^{(A_t^{t})})}{q(\theta_t| \theta_{t-1}^{(A_t^{t})}, y_t, x_t)}$ \Comment{Compute weights.}
\State $W_t^{(i)} = w_t^{(i)} / \sum_{k=1}^N w_t^{(k)}$ \Comment{Normalize weights.}
\EndFor
\end{algorithmic}
\end{algorithm}

The main design element in Sequential Monte Carlo is to construct the proposal distributions. 
A convenient, and common, approach is to sample directly from the transition density, 
$q(\theta_t| \theta_{t-1}, y_t, x_t) = p(\theta_t | \theta_{t-1})$.
This is known as the \textit{Bootstrap filter} \citep{chopin2020introduction}. 
While easy to implement, the Bootstrap proposal does not consider the observed data. 
This can lead to sample depletion if the likelihood is very informative \citep{doucet2001sequential}.
One can alleviate this by increasing the noise level in the observations, but this introduces an additional tuning parameter.
Guided proposals attempt to generate samples that take the data into account, 
$q(\theta_t| \theta_{t-1}, y_t, x_t) \approx p(y_t | \theta_t, x_t) p(\theta_t | \theta_{t-1})$,
which hopefully leads to a higher sample efficiency.
These proposals can, however, be challenging to design.

The measure of sample efficiency that will be used in this study is the \textit{effective sample size}.
It attempts to capture how many particles were obtained in practice when the sample importance is taken into account.
The effective sample size at time $t$ is calculated as $\text{ESS}(W_t^{1:N}) = 1/(\sum_{i=1}^N (W_t^{(i)})^2)$, 
and takes values between 1 and $N$ \citep{chopin2020introduction}.
We will report the effective sample size divided by the number of samples.

\section{Problem description} \label{sec:problem}
We consider a set of $m$ wells that produce to a shared separator facility as illustrated in Figure \ref{fig:asset}.
The wells are instrumented as illustrated in Figure \ref{fig:well}.
For well $j$ at time $t$, the measurements are collected in a vector denoted $x_{j,t} \in \mathbf{R}^4$.
These contain choke opening, pressures, and temperature.
All wells produce a mixture of gas, oil, and water.
The flow rate from a well is denoted 
$y_{j,t}^\top = \begin{bmatrix} y_{j,t,G} & y_{j,t,O} & y_{j,t,W} \end{bmatrix}$.
Let $y_{j,t,T} = y_{j,t,G} + y_{j,t,O} + y_{j,t,W}$ be the total flow from the well,
and $\phi_{j,t} = y_{j,t}/y_{j,t,T}$ be the flow composition.

We consider VFMs that are only capable of estimating the total flow through the choke and require knowledge about the flow composition.
They take the form $y_{j,t,T} = f_j(x_{j,t}, \phi_{j,t}) + v_{j,t}$, with an error term $v_{j,t}$.
Our goal is to infer the unknown composition $\phi_{j,t}$, which is required by the VFMs.
Additionally, the VFMs require calibration in form of a tuning factor that scales the predicted values. 
Let $\beta_{j,t} \in \mathbf{R}$ be the tuning factor for well $j$ at time $t$.
The multiphase flow rates are then given by
\begin{align}
y_{j,t} &= \beta_{j,t} \phi_{j,t} f_j(x_{j,t}, \phi_{j,t}) + e_{j,t}, \label{eq:model-estimate} \\
e_{j,t} &\sim \mathcal{N}\left(0, \Sigma_{j,t} \right). \label{eq:model-estimate-e}
\end{align}
The VFM itself is a known function $f_j:\mathbb{R}^4 \times [0, 1]^3 \mapsto\mathbb{R}$, that can be evaluated freely given the measurements and composition.
Further details about the functions $f_j$ are given in Appendix \ref{app:model}.

To perform the inference, we have access to measurements from the separator.
The flow rates measured at the separator, $y_t$, is the sum of production from all wells, 
\begin{align}
   y_t &= \sum_{j=1}^{m} y_{j,t} + \epsilon_t, \label{eq:mass-balance-y}
\end{align}
where $\epsilon_t \sim \mathcal{N}(0, \Sigma_{\epsilon,t})$.
From each well, we have the measurements $x_{j,t}$.
The complete data set is given as $\mathcal{D}_{1:n} = \lbrace (y_t, X_{t}) \rbrace_{t=1}^{n}$,
where $X_t = \lbrace x_{1,t}, \dots, x_{m,t}  \rbrace$.
If for some time step $t$, the flow rates from all but one well are zero, the data point is a well test. 
Otherwise, it is a production data point.

The goal is to calibrate the VFMs.
This means to provide estimates of the tuning factor $\beta_{j,t}$ and flow composition $\phi_{j,t}$ for all wells, based on data seen up to time step $t$.
The composition vector $\phi_{j,t}$ have two degrees of freedom 
and can be parametrized by the gas fraction 
\begin{align}
    \gamma_{j,t} = \frac{y_{j,t,G}}{y_{j,t,G} + y_{j,t,O} + y_{j,t,W}} \label{eq:gamma}
\end{align}
and the oil factor 
\begin{align}
    \lambda_{j,t} = \frac{y_{j,t,O}}{y_{j,t,O} + y_{j,t,W}}. \label{eq:lambda}
\end{align}
where both $\gamma_{j,t}$ and $\lambda_{j,t}$ take values in the unit interval.
These are related to the composition by
\begin{align*}
   \phi_{j,t} &= \begin{bmatrix} 
   \gamma_{j,t} \\
   (1-\gamma_{j,t}) \lambda_{j,t} \\
   (1-\gamma_{j,t}) (1-\lambda_{j,t})
   \end{bmatrix},
\end{align*} 
The set of parameters for well $j$ at time $t$ is denoted
$\Theta_{j, t} = \lbrace \beta_{j,t}, \gamma_{j,t}, \lambda_{j,t}, \rbrace$,
and the full set of parameters at time $t$ is $\Theta_{t} = \lbrace \Theta_{1,t}, \dots, \Theta_{m,t}, \rbrace$.

\subsection{State space model} \label{sec:model}

The filtering problem is to sample from $p(\Theta_{t} | \mathcal{D}_{1:t})$, for $t=1, \dots, n$. 
To apply the Sequential Monte Carlo framework to this problem we need to specify the priors, state transition densities, and likelihood of the observations.

For the state transitions,
both the composition and tuning factors are assumed to be Markov chains that are mostly constant,
but with a small probability of making a change.
An augmentation to the parameters, $Z_t = \lbrace z_{1,t}, \dots, z_{m,t} \rbrace$, are introduced to model these step changes.
A change occurring in well $j$ at time $t$ is given by 
\begin{align}
    z_{j,t} \sim \text{Bernoulli}(p_z). \label{eq:jump}
\end{align}
One variable is used for each well, which means that either all or none of the three well parameters are able to change at any given time.
If there is a change, the transitions are modeled by truncated Normal distributions.
Let $\mathcal{N}_a^b$ denote a Normal distribution truncated to the interval $[a, b]$. 
The special case $\mathcal{N}_0^\infty$ is the half-normal distribution.
The prior and transition equations for the tuning factors are
\begin{align}
\beta_{j,0} &\sim \mathcal{N}_0^\infty(\mu_{\beta,0}, \sigma_{\beta,0}^2), \\
\beta_{j,t} &= (1-z_{j,t})\beta_{j,t-1} + z_{j,t} \tilde\beta_{j,t}, \label{eq:update-with-jump} \\ 
\tilde\beta_{j,t}  &\sim \mathcal{N}_0^\infty(\beta_{j, t-1}, \sigma_{\beta}^2), \label{eq:update-beta}
\end{align}
The same structure is used for the composition parameters, but with truncation to the unit interval,
\begin{align}
\tilde{\gamma}_{j,t} &\sim \mathcal{N}_0^1 (\gamma_{j,t-1}, \sigma_{\gamma}^2), \label{eq:update-gamma} \\
\tilde{\lambda}_{j,t} &\sim \mathcal{N}_0^1(\lambda_{j,t-1}, \sigma_{\lambda}^2). \label{eq:update-lambda}
\end{align}
Equations \eqref{eq:jump}--\eqref{eq:update-lambda} results in the transition density
\begin{align}
    p(\Theta_{t}, Z_t | \Theta_{t-1}, Z_{t-1}) = p(\Theta_{t} | \Theta_{t-1}, Z_t) p(Z_t). \label{eq:transition-density}
\end{align}

The likelihood term, $p(y_t | \Theta_{t}, X_t)$, is constructed by combining Equation \eqref{eq:mass-balance-y} and Equation \eqref{eq:model-estimate},
which yields the expression
\begin{align}
y_t &= \sum_{j=1}^{m} \left[ \beta_{j,t} \phi_{j,t} f_j(x_{j,t}, \phi_{j,t}) + e_{j,t} \right] + \epsilon_{t}. \label{eq:observer-sum}
\end{align}
The likelihood does not depend on the augmented variables introduced in Equation \eqref{eq:transition-density},
and the resulting likelihood is simply given as 
\begin{align}
y_{t} &\sim \mathcal{N}\left(\bar{y}_{t}, \Sigma_t \right), \label{eq:likelihood} \\
\bar{y}_{t} &= \sum_{j=1}^{m} \phi_{j,t}\beta_{j,t} f_j(x_{j,t}, \phi_{j,t}), \label{eq:exp-y} \\
\Sigma_{t} &= \Sigma_{\epsilon,t} + \sum_{j=1}^{m} \Sigma_{j,t}. \label{eq:std-y}
\end{align}
The individual error terms, $\Sigma_{j,t}$, are time-dependent. 
If a well is closed at time $t$, then its error term will not contribute to the observation.

\subsection{Data}

\begin{figure*}
    \centering
    \includegraphics[width=\columnwidth, page=1]{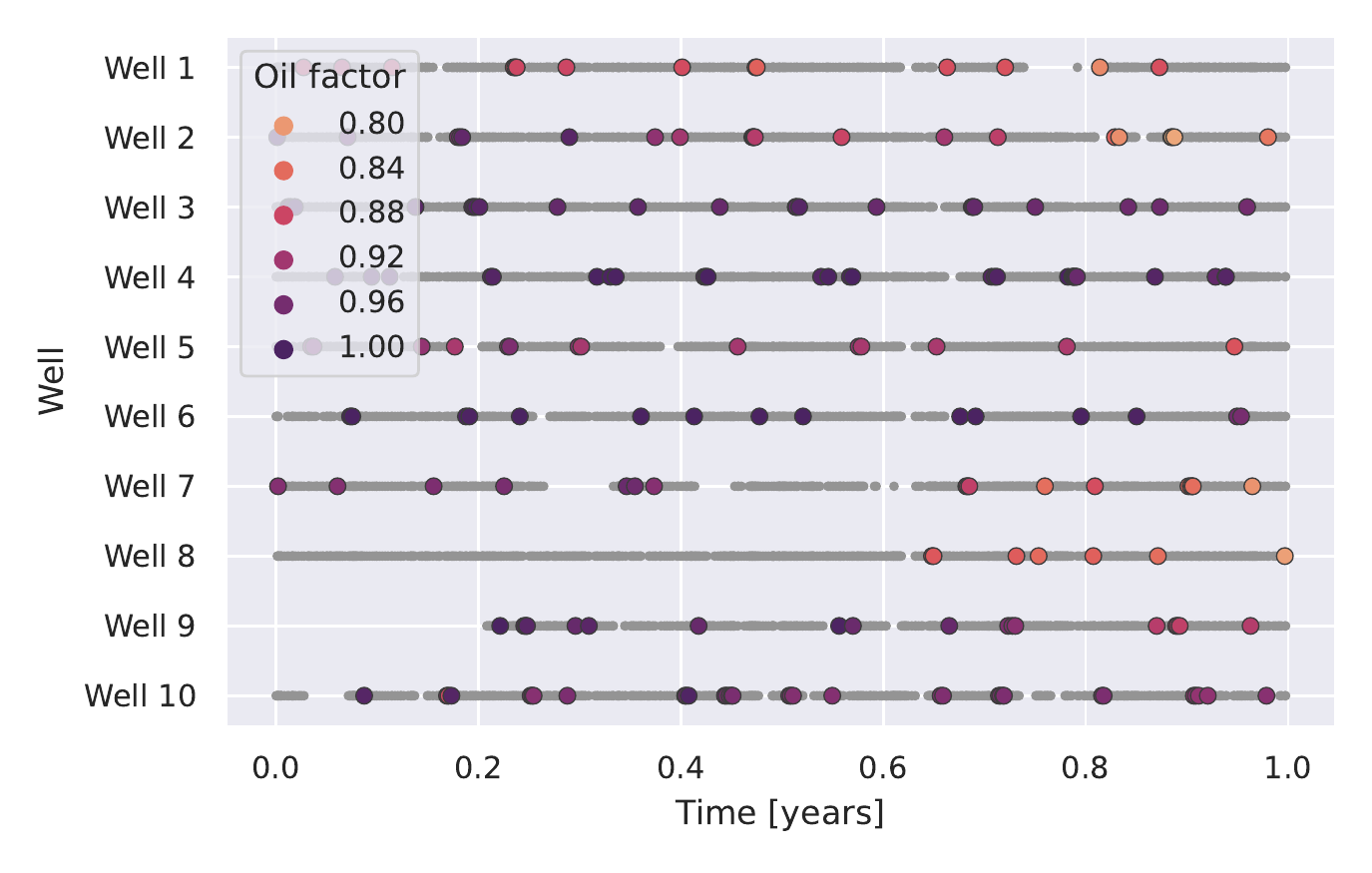} 
    \caption{
    Overview of data.
    Each line corresponds to one well.
    A grey point is plotted for each \textit{production} data point where the well is producing.
    A colored point is plotted for each \textit{well test}, with the color indicating the oil factor computed for that test.
    }
    \label{fig:data-overview}
\end{figure*}

Our dataset is one year of data taken from an asset with ten wells. 
The asset has two separators, one for well testing and one for the main production line.
We are primarily using the production data for inference, and the well test data for performance evaluation.
An overview of how the data is distributed in time is plotted in Figure \ref{fig:data-overview} and the quantities to be estimated are summarized in Table \ref{tab:data-summary}. 
We read from Table \ref{tab:data-summary} that 
We see that some wells are quite constant over the entire period, while others see a continuous change to their oil factor.
The wells are also all quite similar to each other.
Also note that one well is not tested during the first half of the dataset.

\begin{table}
    \caption{Summary of data points. 
    Listed is the number of production data points the well is contributing to, the number of welltests, and the minimum and maximum flow composition values.}
\centering
\begin{tabular}{lrrrrrr}
\hline
         &   Production &   Well test &   min $\gamma_j$ &   max $\gamma_j$ &   min $\lambda_j$ &   max $\lambda_j$ \\
\hline
 Well 1  &          622 &          14 &             0.13 &             0.14 &              0.81 &              0.90 \\
 Well 2  &          664 &          21 &             0.14 &             0.16 &              0.78 &              1.00 \\
 Well 3  &          643 &          22 &             0.14 &             0.17 &              0.96 &              1.00 \\
 Well 4  &          645 &          29 &             0.15 &             0.16 &              0.97 &              1.00 \\
 Well 5  &          635 &          14 &             0.08 &             0.15 &              0.86 &              0.95 \\
 Well 6  &          641 &          15 &             0.15 &             0.17 &              0.96 &              1.00 \\
 Well 7  &          565 &          17 &             0.13 &             0.29 &              0.81 &              0.97 \\
 Well 8  &          666 &           7 &             0.05 &             0.14 &              0.79 &              0.87 \\
 Well 9  &          492 &          18 &             0.14 &             0.15 &              0.89 &              1.00 \\
 Well 10 &          536 &          33 &             0.14 &             0.16 &              0.87 &              0.99 \\
\hline
\end{tabular}
\label{tab:data-summary}
\end{table}

\section{Results and discussion} \label{sec:results}

We study the feasibility of inferring the latent state-space variable using SMC by exploring three variations of our dataset.
In addition to the real data,
two \textit{semi-synthetic} cases are explored to highlight the properties of the method under controlled circumstances. 
These cases are constructed using real data as a basis.
Two variations of the real dataset are considered.
One where \textit{only} production data is used for inference and well tests are only used for validating the performance,
and one where \textit{both} production data and well test data are used for inference.

Validation is done with values derived from the well tests as the target and the sample average from the \textit{previous} time step as the estimates.
The targets for the gas fraction and oil factors are computed from the well tests according to Equation \ref{eq:gamma} and Equation \ref{eq:lambda}.
The target for the tuning factor is found by solving Equation \ref{eq:model-estimate} for $\beta_{j,t}$.
For real data, the target is taken as the well tests. 
For synthetic data, the target is the underlying true values.
All visualizations of the results are done using the sample mean, along with colored bands that cover the 25th to 75th percentile and the 5th to 95th percentile in two different shades.

In all cases, we use the bootstrap filter to produce the proposals.
This is straightforward according to the description in Section \ref{sec:model}.
In all cases we run the algorithm with 100 000 samples.
The specific parameters used for the models are given in Appendix \ref{app:parameters}.

\subsection{Constructed three well case} \label{sec:case-three-well}

\begin{figure*}
    \centering
    \includegraphics[width=\columnwidth, page=1]{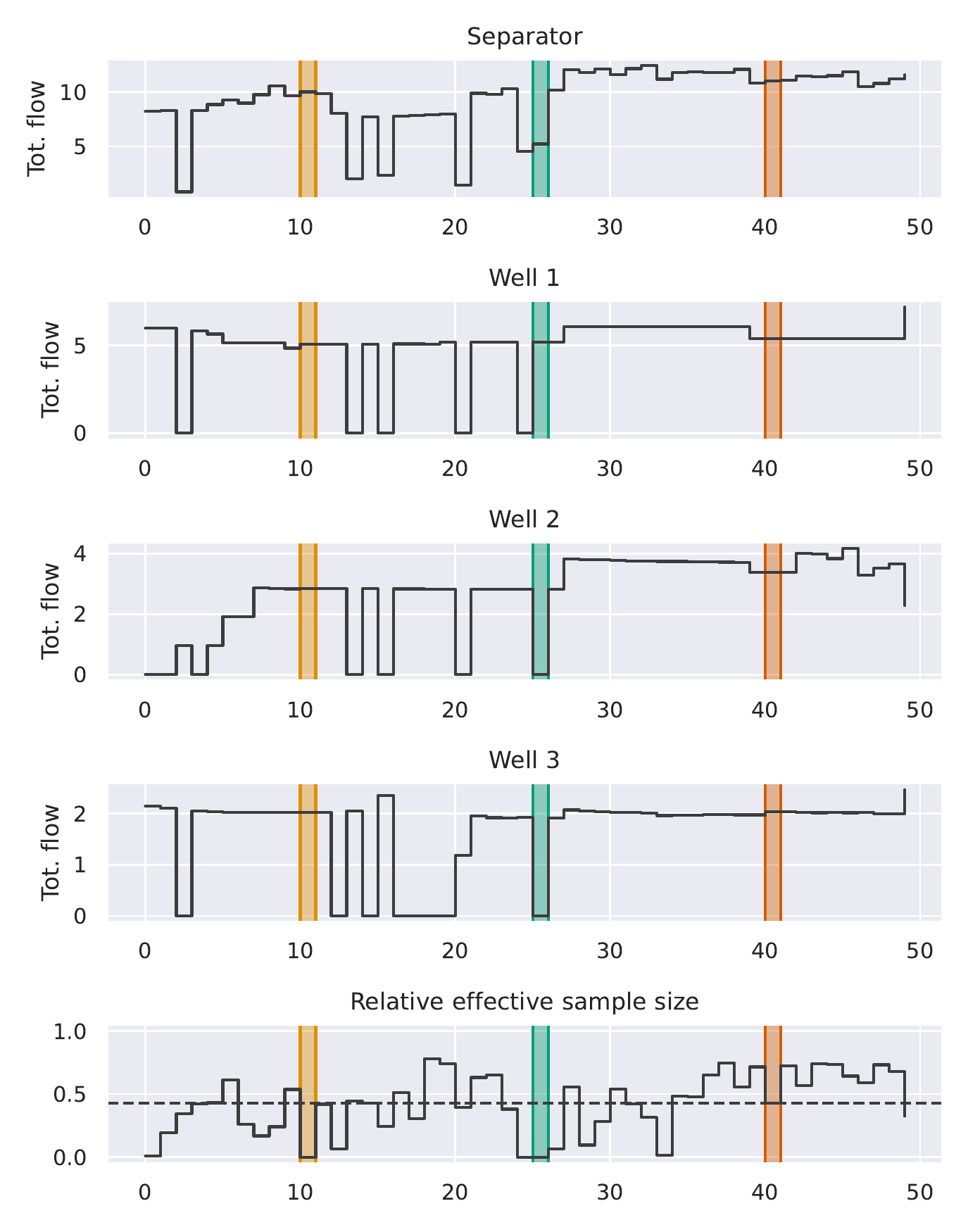} 
    \caption{
    Total flow rates for the constructed case with $m=3$. 
    The top plot is the sum of separator flow rates, which is measured. 
    The three middle plots are the total flow rates from the three wells.
    The bottom plot is the relative effective sample size from the SMC estimates.
    The oil factor samples are illustrated in Figure \ref{fig:simple-samples}, for the three highlighted time steps.
    }
    \label{fig:simple-case-qtot}
\end{figure*}

We begin by investigating a constructed case with three wells.
Only the first 50 data points, including well tests, are considered.
All state parameters are set to constant values, except for a jump in oil factor for Well 2 at time step $t=10$.
The features, $x_{j,t}$ are taken to be equal to those from the real data,
but flow rates are computed as the sum of the VFMs evaluated with the constructed state values.
Noise is added to the separator measurement according to the model.
This means that the case has no modeling errors because the VFMs and the state-space model perfectly match the system.
The resulting total flow rates are illustrated in Figure \ref{fig:simple-case-qtot}.

\begin{figure*}
    \centering
    \includegraphics[width=\columnwidth, page=4]{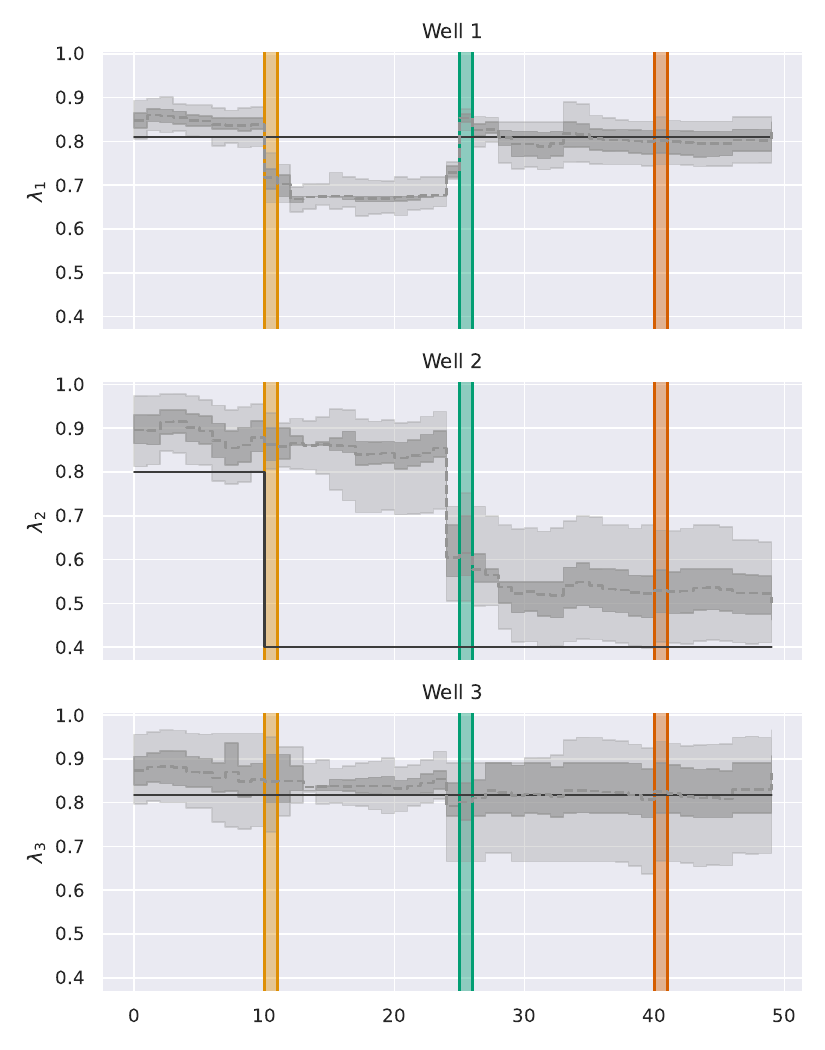} 
    \caption{
    The oil factors estimated from the data given in Figure \ref{fig:simple-case-qtot}.
    Shown are the marginal sample distributions at each time step in grey and the truth in black.
    The colored intervals are investigated in Figure \ref{fig:simple-lambda}.
    }
    \label{fig:simple-lambda}
\end{figure*}

\begin{figure*}
    \centering
    \includegraphics[width=\columnwidth]{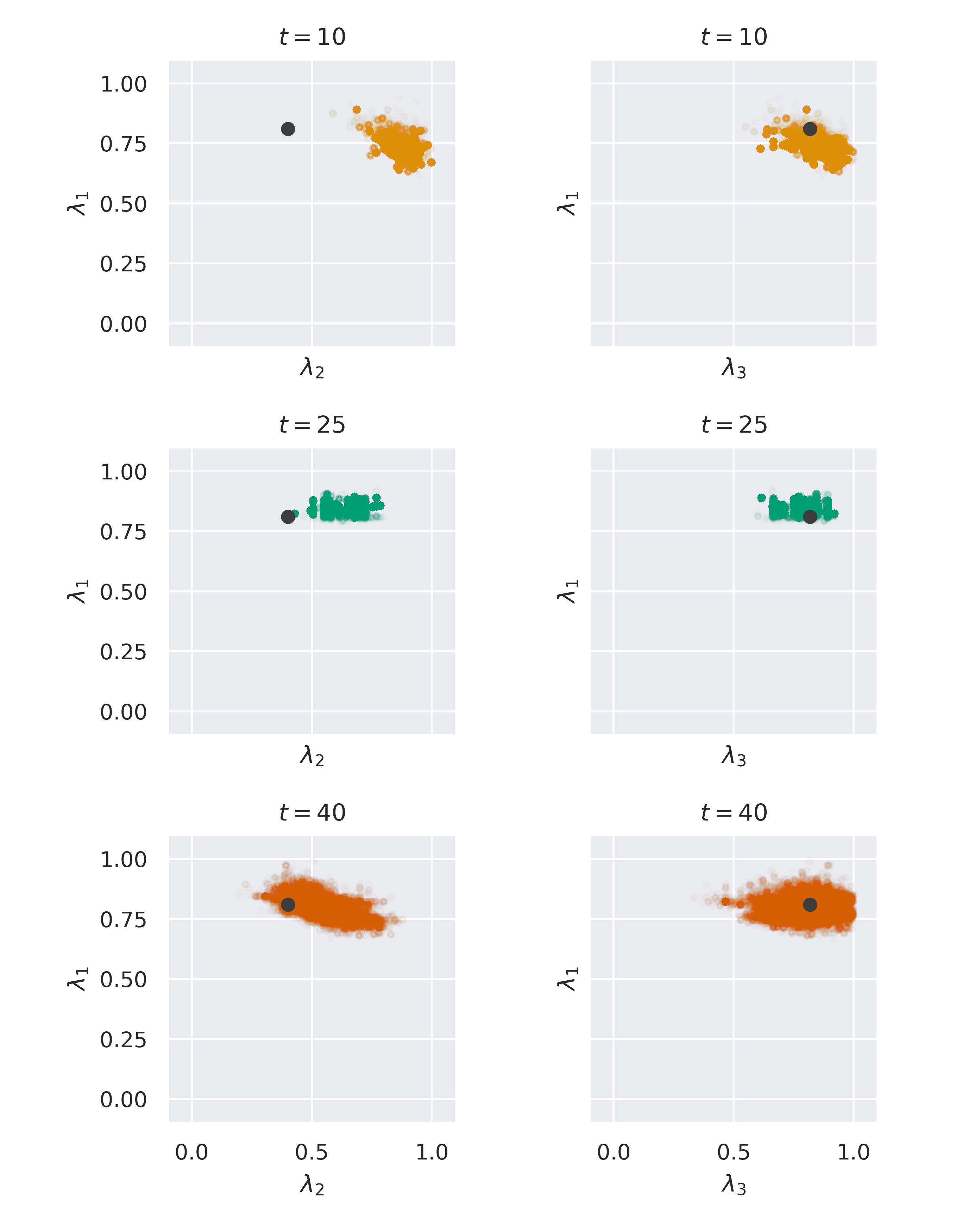} 
    \caption{
    Scatter plot of the oil factor samples, taken at the time steps highlighted in Figure \ref{fig:simple-lambda}.
    The colored points are the samples and the black dot is the true value.
    }
    \label{fig:simple-samples}
\end{figure*}

For this constructed case, we only study the results visually.
The resulting oil factor estimates are illustrated in Figure \ref{fig:simple-lambda}.
In Figure \ref{fig:simple-samples} the samples from the regions highlighted in Figure \ref{fig:simple-case-qtot} and Figure \ref{fig:simple-lambda} are visualized as scatter plots.

As seen in Figure \ref{fig:simple-case-qtot}, 
the effective sample size is mostly healthy, with an average value of slightly below $50\%$.
However, some events cause the effective sample size to drop significantly.
This is seen at $t=10$ when the oil factor of Well 2 changes dramatically.
Other informative events, such as well tests, also cause drops in the effective sample size.
Additionally, during the first few time steps the sample size needs some time to settle.
We see the lack of sample diversity clearly in Figure \ref{fig:simple-samples} for $t=25$.
The sample size does however evolve back to a healthy distribution, illustrated at $t=40$.

The simple case with a single jump in the oil factor reveals an important aspect of our inference problem.
With only the sum of flow rates to inform us about the hidden variables, it is difficult to distinguish where the change has happened.
In this case, the change is, erroneously, assigned to Well 1. 
This remains until informative events happen at $t=24$ and $t=25$, when Well 1 is removed from production and placed on the test separator.
This provides information about the true state of Well 1, which in turn allows for a correction in the other wells too.
It is interesting to note that for Well 2, there is just as much information being provided when Well 1 is \textit{removed} at $t=24$,
while for Well 1 the main correction happens at $t=25$.
The remainder of the sequence lacks the excitation needed to make any further improvements to the estimates.
This reflects how important it is to have variability in this kind of data,
and gives us an indication that time steps where wells are closed or started are the most valuable for inference.

\subsection{Full synthetic case} \label{sec:case-synt}

This case is a synthetic case following the same strategy as presented in Section \ref{sec:case-three-well},
except that all ten wells are included and that the parameters are generated differently.
We present \textit{two variations} on this case. They differ only in how the composition parameters are generated.

In the first case, referred to as \textit{synthetic copy},
the composition is taken as \textit{linear interpolation} of the true well test values.
The tuning factors are all constant and equal to one for all wells.
This is used, along with the real measurements $x_{j,t}$, to evaluate the VFMs to produce the response variables $y_t$.
This case then reflects the real cases, if the VFMs were perfect.
Note that our state-space model assumes the state evolves as a piecewise constant sequence,
while the generated composition used here evolves as a piece-wise linear trend between well tests.
The priors for the parameters are set close to the separator average at the beginning of the data set.

The second case, referred to as \textit{synthetic random},
is identical to the synthetic copy case, except that the compositions do not follow the well tests exactly.
The compositions from each well are assigned a random starting point in the unit interval.
Then, the development over time is set to a shuffled version of the increments seen in the true well tests.
The motivation for this second case is that the compositions seen in the real data are all very similar,
which makes the asset average a very good estimate of the individual wells.

\begin{figure*}
    \centering
    \includegraphics[width=\columnwidth, page=1]{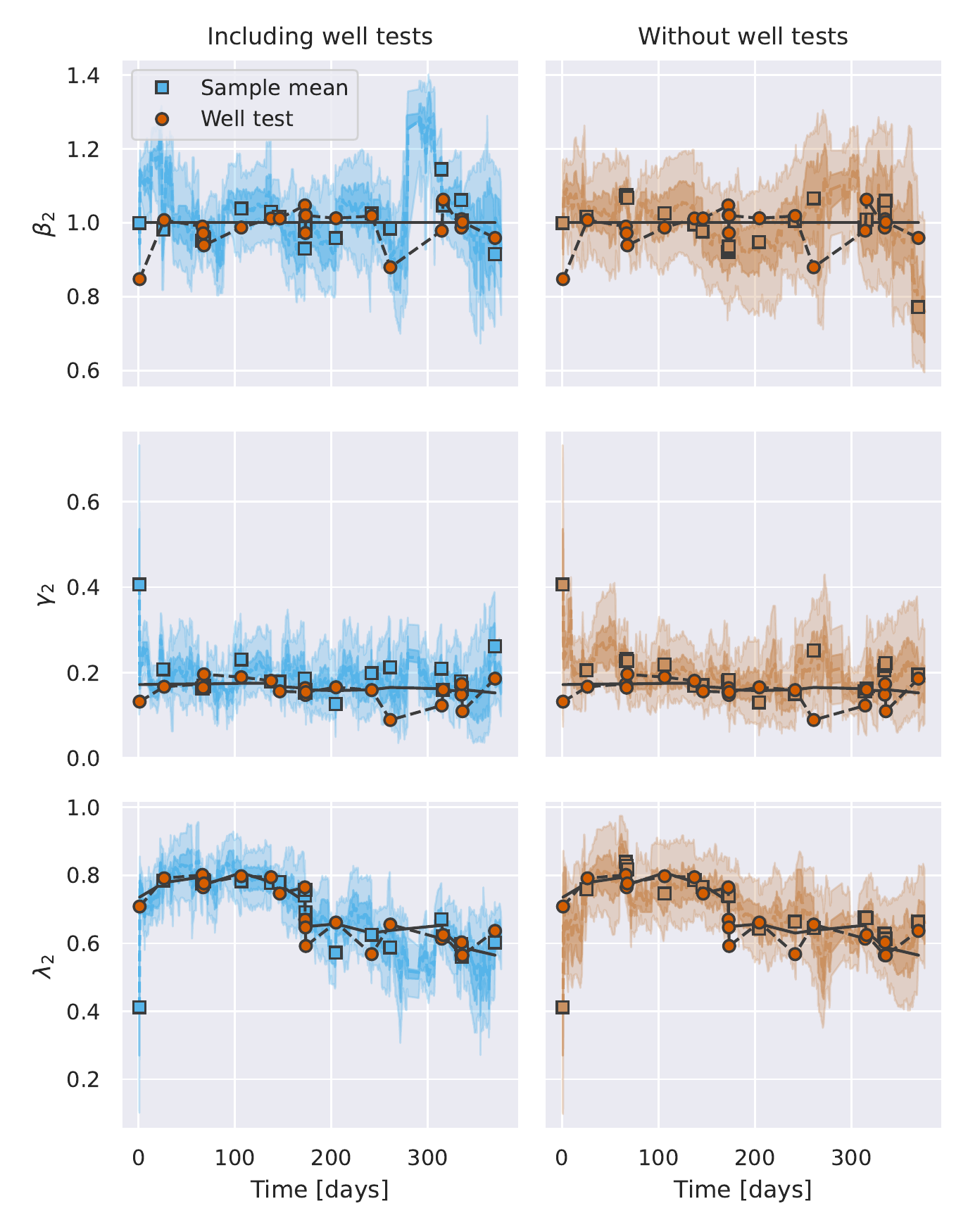} 
    \caption{
    Marginal estimates for Well 2 for the synthetic random.
    The left column is the estimates where well tests are included in the data set,
    and the right column is the estimates without well tests.
    The rows are the tuning factor, gas fraction, and oil factor respectively.
    The orange dots are the parameters computed from the well test values.
    The blue and brown squares are the sample mean at the time step \textit{before} the well test.
    }
    \label{fig:synt-well-2}
\end{figure*}

\begin{figure*}
    \centering
    \includegraphics[width=\columnwidth, page=1]{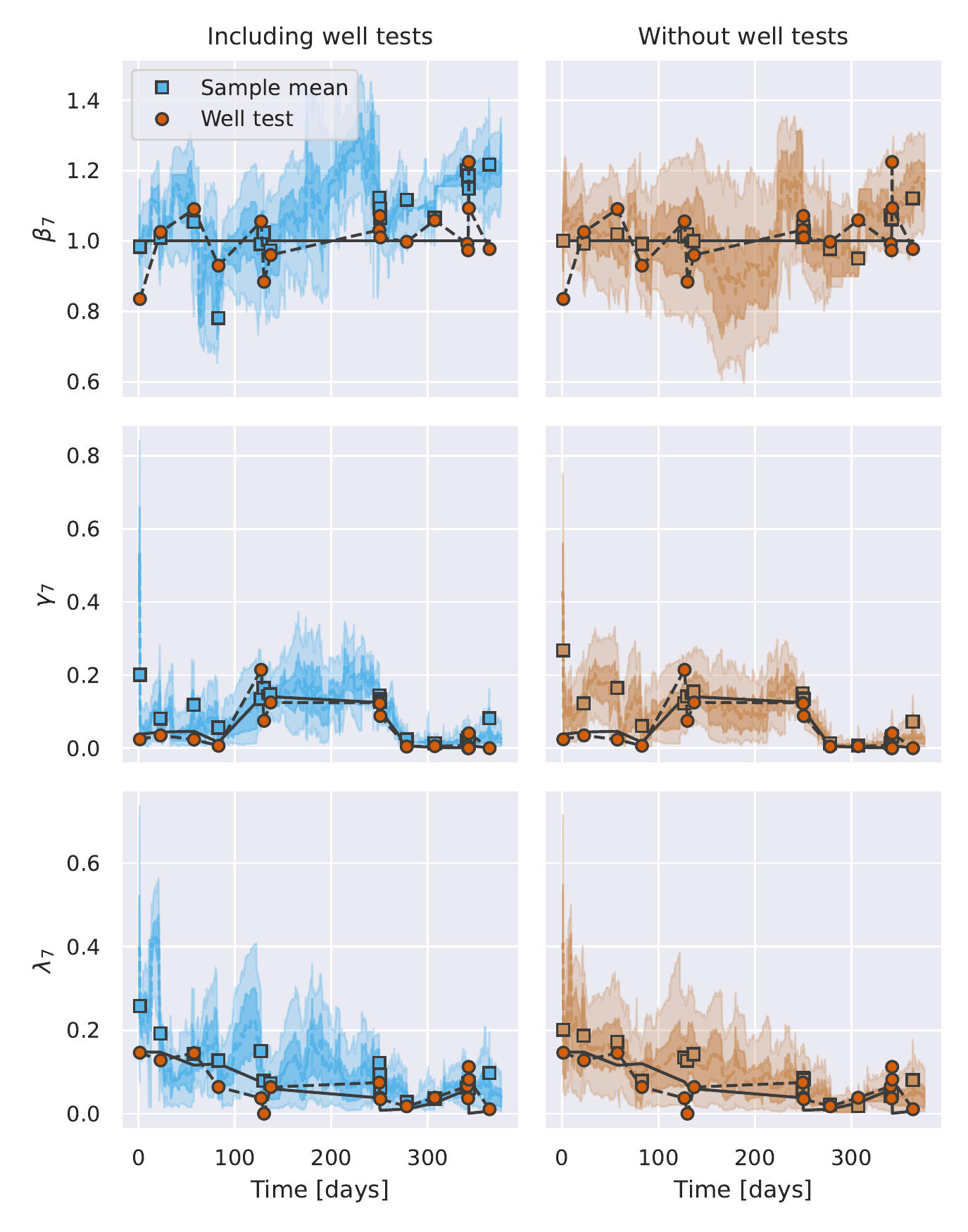} 
    \caption{
    Marginal estimates for Well 7 for the synthetic random case.
    The content is the same as in Figure \ref{fig:synt-well-2}.
    }
    \label{fig:synt-well-7}
\end{figure*}

\begin{figure*}
    \centering
    \includegraphics[width=\columnwidth, page=1]{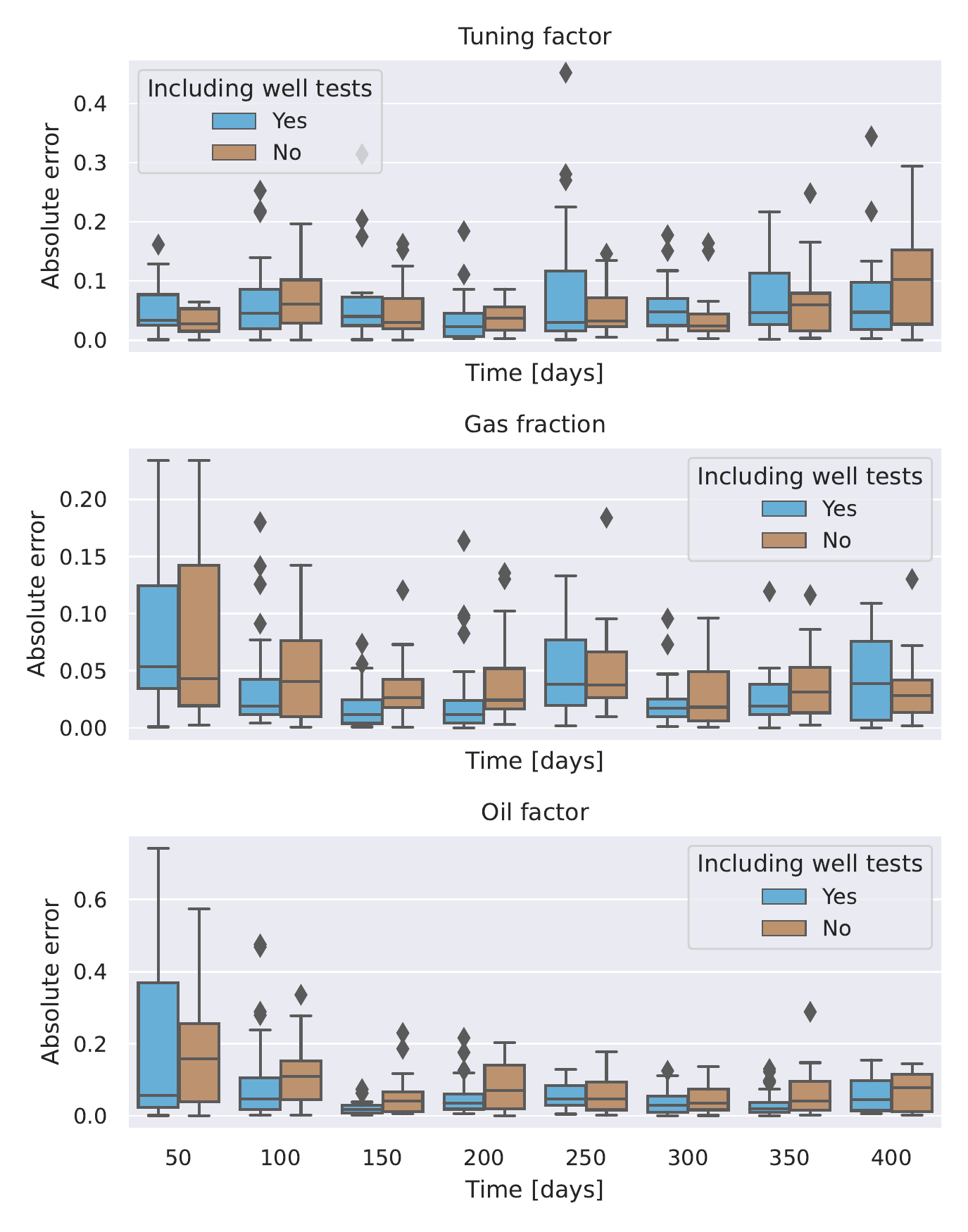} 
    \caption{
    Box plot of the absolute error on the synthetic random case,
    grouped into intervals of 50 days.
    }
    \label{fig:synt-box-plot}
\end{figure*}

We study the synthetic random case in detail.
The marginal estimates for Well 2 and Well 7 are given in Figure \ref{fig:synt-well-2} and Figure \ref{fig:synt-well-7} respectively.
For both wells we see the first days are heavily influenced by the prior, which is close to 0.4 for both gas fraction and oil factor in this case.
The estimates eventually converge towards the correct values, with the inference that has access to the well tests naturally converging slightly faster.
We see the same effect when the estimation errors are aggregated across all wells in Figure \ref{fig:synt-box-plot}.
The estimates are able to follow the developments of the composition quite well, even in the case without well tests.
For both wells the tuning factor estimates are quite stable, but a slight drift is occasionally observed in both cases.

The average performance on both synthetic cases is summarized in Table \ref{tab:mad-summary}.
We see that the effects of including well tests are higher in the random case than in the copy case.
In the copy case, this is suspected to be because the compositions are all quite similar, and the prior is therefore quite powerful.
Additionally, simply setting the composition equal to the asset average is going to be a good estimate on its own.
Another reason for the close performance on both cases is that the information provided by \textit{removing} a well from production can be quite significant, as was noted in Section \ref{sec:case-three-well}.
We finally note that the average sample size is acceptable for both datasets.

\subsection{Real data case}\label{sec:case-real}

Finally, we conduct the inference on the full real dataset.
In this case, the VFMs are not perfectly accurate, 
and there is no guarantee that the values observed on the test separator translate directly to the values seen on the production separator.
This also means that the values used for the noise terms do not describe the truth.

\begin{figure*}
    \centering
    \includegraphics[width=\columnwidth, page=1]{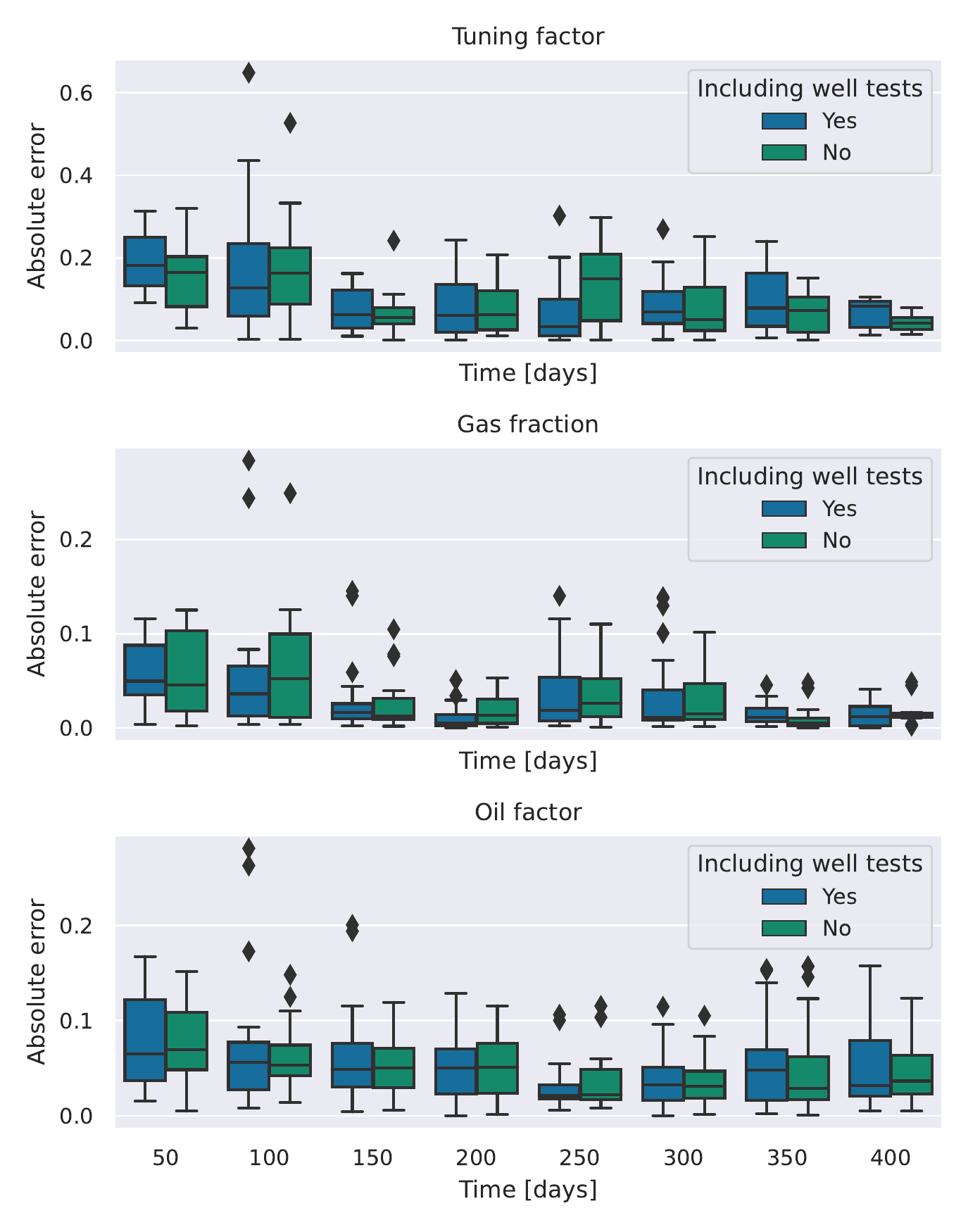} 
    \caption{
    Box plot of the absolute error on real data, grouped into intervals of 50 days.
    }
    \label{fig:real-box-plot}
\end{figure*}

The average performance is summarized in Table \ref{tab:mad-summary} along with the synthetic data results,
and the development of the errors over time is visualized in Figure \ref{fig:real-box-plot}.
In both cases, the estimates require approximately 100 days of data for the samples to settle and for the errors to stabilize.
The errors are significantly higher with real data, compared to the synthetic copy case.
This is natural, because the VFMs are quite simple models, and are unlikely to capture the full operating range seen over a year. 
This means that the tuning factors take an active role, and they can have large changes from one test to the next.
this also leads to quite high errors. 
For instance, a tuning factor error of $0.1$ corresponds to a 10\% error in the multiplicative correction factors.

There is little difference between the errors observed with and without well tests included in the dataset.
This was also observed with the synthetic copy case, and possible reasons were discussed in Section \ref{sec:case-synt}.

\begin{figure*}
    \centering
    \includegraphics[width=\columnwidth, page=1]{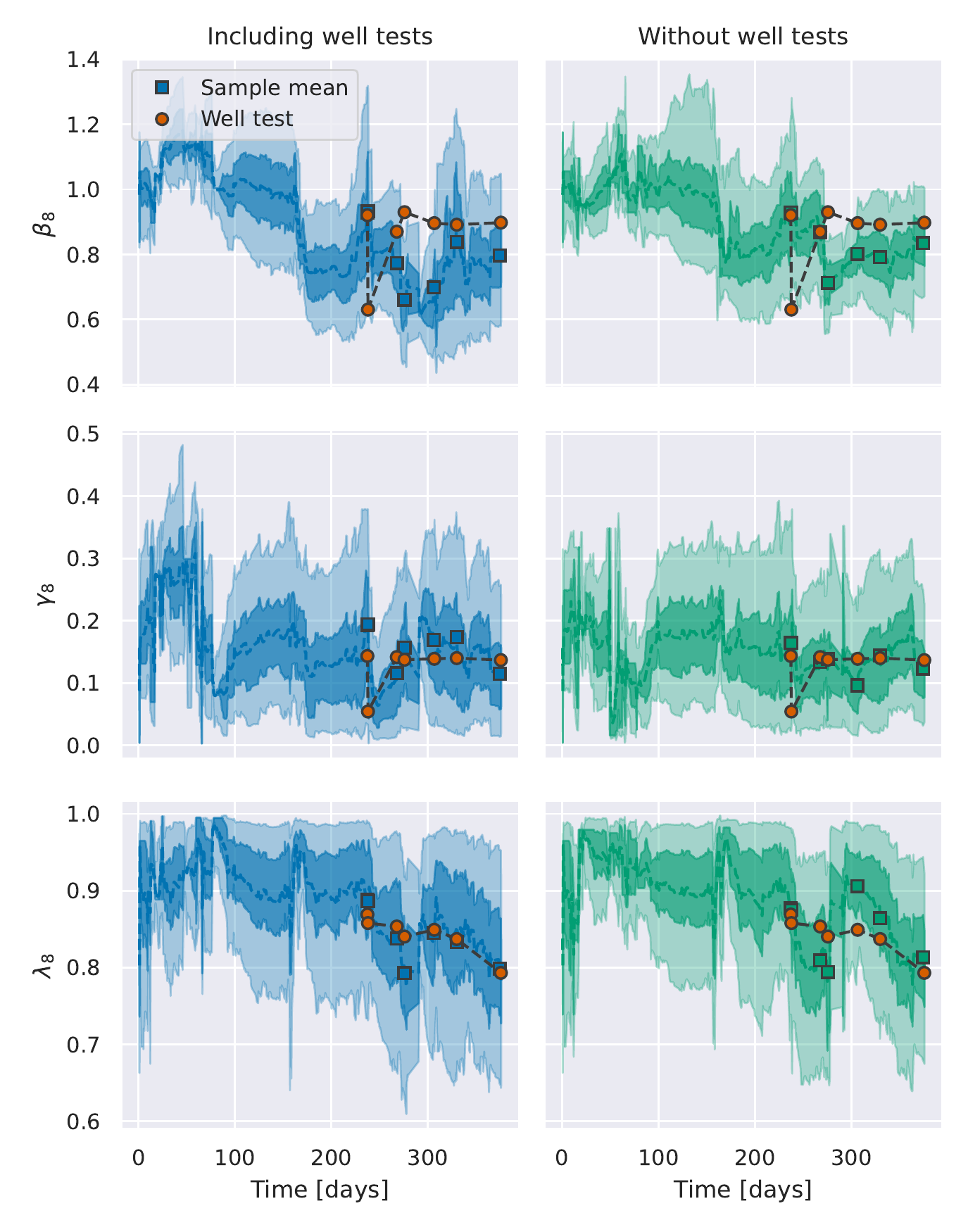} 
    \caption{
    Marginal estimates for Well 8 for the real data set.
    The left column shows the estimates where well tests are included in the data set,
    and the right column shows the estimates without well tests.
    The rows are the tuning factor, gas fraction, and oil factor respectively.
    The orange dots are the parameters computed from the well test values.
    The blue and green squares are the sample mean at the time step \textit{before} the well test.   
    }
    \label{fig:real-well-8}
\end{figure*}

Figure \ref{fig:real-well-8} illustrates the estimates for Well 8, which has its first well test after approximately 220 days.
The inference deduced reasonable values for its parameters based solely on the joint observations from the production data.
It can be noted that the parameter values computed from the well tests are quite close to the prior assumption. 
However, both the gas fraction and the oil factor estimates take on different values during the first 200 days,
before settling back to the values observed in the last half of the time series,
which indicates that the accuracy of the estimate is not only due to the prior.

\begin{figure*}
    \centering
    \includegraphics[width=\columnwidth, page=1]{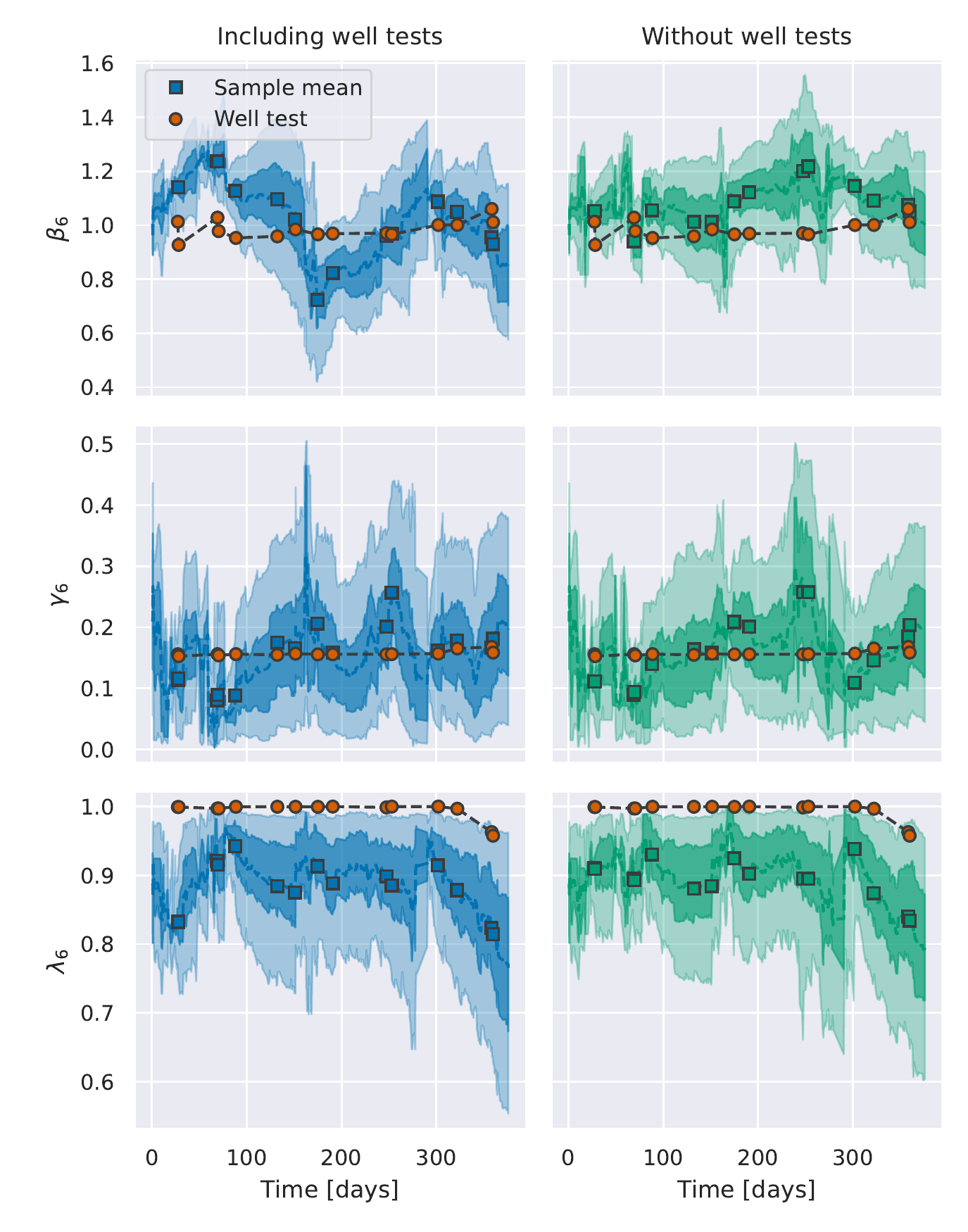} 
    \caption{
    Marginal estimates for Well 6 for the real data set.
    Figure content is similar to Figure \ref{fig:real-well-8}.
    }
    \label{fig:real-well-6}
\end{figure*}

The noise levels are not as accurately calibrated in the real case, as for the synthetic ones.
A consequence of this is seen in Figure \ref{fig:real-well-6}.
The oil factor of Well 6 is close to one, 
but due to uncertainty, the sample mean is biased away from the boundary. 
In the case of Well 6, this means that the sample mean for the oil factor remains in the range of 0.90, no matter how many data points are collected.
We could adjust the noise levels down, but this can also be problematic.
For instance, consider the gas fraction in Figure \ref{fig:real-well-8}.
Of the seven well tests, one has a significantly lower gas fraction than the others.
This may be due to some additional process disturbance that is difficult to capture with a static Normal noise model.
The result is a trade-off where most well tests are being given too little influence, because otherwise, the outliers would have a too large impact on the importance weights.
In practice, each test could be assessed and assigned individual noise levels.
This would lead to highly subjective priors and is not pursued here.

\begin{table}
\caption{Summary of performance on the three cases, real, synthetic copy, and synthetic random. 
All cases is listed with and without well tests included in the dataset.
    In all cases the absolute errors are averaged over the ten wells, which is 190 well tests combined.}
\centering
\begin{tabular}{lrrrrrr}
\hline
Data type           &   Real    &  Real     &  S. copy  &   S. copy &   S. random   &   S. random \\
Welltest  &   Yes     &  No       &  Yes          &   No          &   Yes             &   No \\
\hline
 Tuning factor      &   0.104   &  0.100    &  0.055        &   0.056       &   0.063 &            0.053 \\
 Gas fraction       &   0.031   &  0.031    &  0.023        &   0.025       &   0.036 &            0.043 \\
 Oil factor         &   0.055   &  0.052    &  0.034        &   0.038       &   0.066 &            0.081 \\
\hline
 Rel. ESS           &   0.598   &  0.531    &  0.358        &   0.385       &   0.396 &            0.417 \\
\hline
\end{tabular}
\label{tab:mad-summary}
\end{table}

\section{Conclusion} \label{sec:conclusion}

A VFM calibration approach based on SMC was presented. The method relied on steady-state data from the production separator, combined with a mass balance, to update tuning factors and flow composition parameters for ten VFMs simultaneously.
The method was investigated on both real and synthetic data with promising results.
The inclusion of well test data is observed to be beneficial, but the method can achieve decent performance also in cases where well tests are lacking.
However, the performance is highly dependent on the levels of excitation seen in the production data, 
which will vary on a case-by-case basis.
It is challenging to provide a fixed set of parameters that capture the process and measurement noise, due to the possibility of outliers. 
This leads to a situation where highly informative events, such as well tests, are given limited influence.
In practice, it would be recommended to have a more complex method of assessing the accuracy of individual data points. 

\appendix

\section{Virtual flow meter specification} \label{app:model}
This section specifies the details of the model and virtual flow meter functions presented in  Section \ref{sec:model}. 

We use the mechanistic flow model of \citet{Sachdeva1986} for our virtual flow meters. 
The particular formulation used is presented in \citet{Hotvedt2022}.
The measured components of $x_{j,t}$ are denoted 
$u$ for the choke opening,
$p_1$ for the upstream choke pressure,
$p_2$ for the downstream choke pressure, 
and $T$ for the temperature. 
The subscripts are ignored for convenience.
It finds the total mass flow as
\begin{align}
y_T &= C_D A(u) \sqrt{
        2\rho_2^2 p_1 \left[
        \frac{\kappa}{\kappa - 1} \phi_G \left(\frac{1}{\rho}_{G,1} - \frac{p_r}{\rho_{G,2}}\right) +
            \left(\frac{\phi_O}{\rho_O} + \frac{\phi_W}{\rho_W} \right) \left( 1-p_r \right)
            \right]
            }. \label{eq:sachdeva}
\end{align}

The model distinguishes between critical and sub-critical flow through the pressure ratio term
$p_r = \max(p_2 / p_1, r_{c})$.
Because $p_2 \leq p_1$, the ratio is a value between zero and one, with the critical pressure ratio, $r_c$, usually being close to 0.6.
The gas densities at the two sides of the choke is given by 
\begin{align*}
    \rho_{G,1} &= \frac{p_1 M}{Z R T_1}, \\
    \rho_{G,2} &= \rho_{G,1} p_r^{1/\kappa},
\end{align*}
where $\kappa$ is the gas expansion coefficient, $Z$ the compressibility factor, $M_G$ the molar mass, and $R$ the ideal gas constant.
This attempts to capture effects related to gas compressibility.
The mixture density downstream is given by a weighted harmonic mean,
\begin{align*}
    \frac{1}{\rho_{2}} = \frac{\phi_G}{\rho_{G}} + \frac{\phi_O}{\rho_{O}} + \frac{\phi_W}{\rho_{W}}.
\end{align*}
The choke geometry is described by the discharge coefficient $C_D$ and an area function $A(u)$.
For additional details about this particular model, see \citet{Hotvedt2022}.

\section{Parameter settings} \label{app:parameters}
This section provides the parameters used in the state space model for all experiments in Section \ref{sec:results}.
The prior and transition parameters are given in Table \ref{tab:settings}.
For the real data case, the flow rates are scaled such that $E[\beta_{j,t}] \approx 1$.
For the compositions, the expected prior values are approximately centered around the production separator average at the start of the dataset.

The noise model has three components, one share element modeling the separator measurement and two components from each well.
The separator measurement noise, $\epsilon_t \sim \mathcal{N}(0, \Sigma_{\epsilon,t})$, is taken to be proportional to the measured value,
\begin{align*}
    \Sigma_{\epsilon,t} = \sigma_\epsilon^2 
    \begin{bmatrix}
    y_{t,G} & 0 & 0 \\
    0 & y_{t,O} & 0 \\
    0 & 0 & y_{t,W} \\
    \end{bmatrix},
\end{align*}
with $\sigma_\epsilon = 0.05$. 
The well component $e_{j,t} \sim \mathcal{N}(0, \Sigma_{j,t})$, is broken into two terms
\begin{align*}
    \Sigma_{j,t} = \sigma_e^2 I_3 + \sigma_f^2 (\beta_{j,t}\phi_{j,t}) (\beta_{j,t}\phi_{j,t})^\top,
\end{align*}
with $\sigma_e = 0.05$ and $\sigma_f = 0.05$. 
The first is a static element capturing process disturbances from the well and the second is a prediction error on the total flow rate from the VFM.
This error is multiplied by the tuning factor and composition, following from Equation \eqref{eq:model-estimate}, which leads to a covariance matrix given by the outer product of these terms.

\begin{table}
    \caption{Settings for the prior and transition densities required by the state space model described in Section \ref{sec:model}.}
\centering
\begin{tabular}{lr}
\hline
Parameter       &   Value \\
\hline
            $p_z$  & 0.1    \\
            $\mu_{\beta,0}$  & 1.0    \\
            $\sigma_{\beta,0}$  & 0.1    \\
            $\sigma_{\beta,0}$  & 0.1    \\
            $\sigma_\beta$  & 0.05    \\
            $\sigma_{\gamma,0}$  & 0.1    \\
            $\sigma_\gamma$  & 0.05    \\
            $\sigma_{\lambda,0}$  & 0.1    \\
            $\sigma_\lambda$  & 0.05    \\
            $\sigma_\lambda$  & 0.05    \\
\hline
\end{tabular}
\label{tab:settings}
\end{table}

\bibliographystyle{apalike}
\bibliography{refs}

\end{document}